\documentclass[11pt]{article}

\usepackage[preprint]{acl}

\usepackage{times}
\usepackage{latexsym}
\usepackage{graphicx}
\usepackage{subcaption}
\usepackage{booktabs} 
\usepackage{amssymb}
\usepackage{amsfonts}
\usepackage{booktabs}   
\usepackage{multirow}   

\usepackage{xcolor} 

\usepackage[T1]{fontenc}

\usepackage[utf8]{inputenc}

\usepackage{microtype}

\usepackage{inconsolata}

\usepackage{graphicx}
\usepackage{amsmath}

\title{SDCD: Structure-Disrupted Contrastive Decoding for Mitigating Hallucinations in Large Vision-Language Models}

\author{
  Yuxuan Xia \\
  University of California, \\
  Santa Barbara \\
  \texttt{yuxuanxia@ucsb.edu} \\\And
  Siheng Wang \\
  University of California, \\
  Santa Barbara \\
  \texttt{siheng@ucsb.edu} \\\And
  Peng Li \\
  University of California, \\
  Santa Barbara \\
  \texttt{lip@ucsb.edu}
}

\begin{document}
\maketitle
\begin{abstract}
Large Vision-Language Models (LVLMs) demonstrate significant progress in multimodal understanding and reasoning, yet object hallucination remains a critical challenge. 
While existing research focuses on mitigating language priors or high-level statistical biases, they often overlook the internal complexities of the visual encoding process. 
We identify that visual statistical bias, arising from the inherent \textit{Bag-of-Patches} behavior of Vision Encoders under weak structural supervision, acts as a contributing factor of object hallucinations. 
Under this bias, models prioritize local texture features within individual patches over holistic geometric structures. 
This tendency may induce spurious visual confidence and result in hallucinations. 
To address this, we introduce a training-free algorithm called Structure-Disrupted Contrastive Decoding (SDCD), which performs contrastive calibration of the output distribution by introducing a shuffled structure-disrupted view. 
By penalizing tokens that maintain high confidence under this structure-less view, SDCD effectively suppresses the texture-driven bias. 
Experimental results demonstrate that SDCD significantly mitigates hallucinations across multiple benchmarks and enhances the overall multimodal capabilities of LVLMs.
\end{abstract}

\section{Introduction}

Although Large Vision-Language Models (LVLMs) have achieved remarkable progress in cross-modal understanding and reasoning tasks \cite{liu2023visualinstructiontuning,bai2023qwenvlversatilevisionlanguagemodel,zhu2023minigpt4enhancingvisionlanguageunderstanding,dai2023instructblipgeneralpurposevisionlanguagemodels}, object hallucination remains a significant bottleneck that constrains their reliability and usability \cite{tong2024eyeswideshutexploring, guan2024hallusionbenchadvanceddiagnosticsuite, zhou2024analyzingmitigatingobjecthallucination, chen2024halcobjecthallucinationreduction}. Existing studies commonly attribute hallucinations to language priors within the language model \cite{leng2023mitigatingobjecthallucinationslarge,liu2024payingattentionimagetrainingfree,liu2024mitigatinghallucinationlargemultimodal,manevich2024mitigatinghallucinationslargevisionlanguage}, meaning that during generation the model tends to follow linguistic statistical co-occurrence patterns rather than strictly relying on visual evidence. While this explanation provides important insights into the hallucination phenomenon, its analytical perspective largely remains at the level of language priors and overlooks the internal pathology of the visual encoding process and its causal role in triggering hallucinations.

We show that the emergence of object hallucination is closely related to the internal mechanisms of the Vision Encoder. Due to the inherent Bag-of-Patches behavior of Vision Encoders, LVLMs often infer objects by aggregating \textit{local statistical texture cues within individual patches}, such as fur patterns or repetitive textures, while lacking explicit modeling of global geometric structure across patches. Lacking explicit structural constraints, this texture-dominant, structure-blind perceptual tendency induces inappropriate visual certainty, forming a bias heavily dominated by local patch-level statistics and ultimately leading to hallucinations.

\begin{figure}[t]
    \centering
    \includegraphics[width=\linewidth]{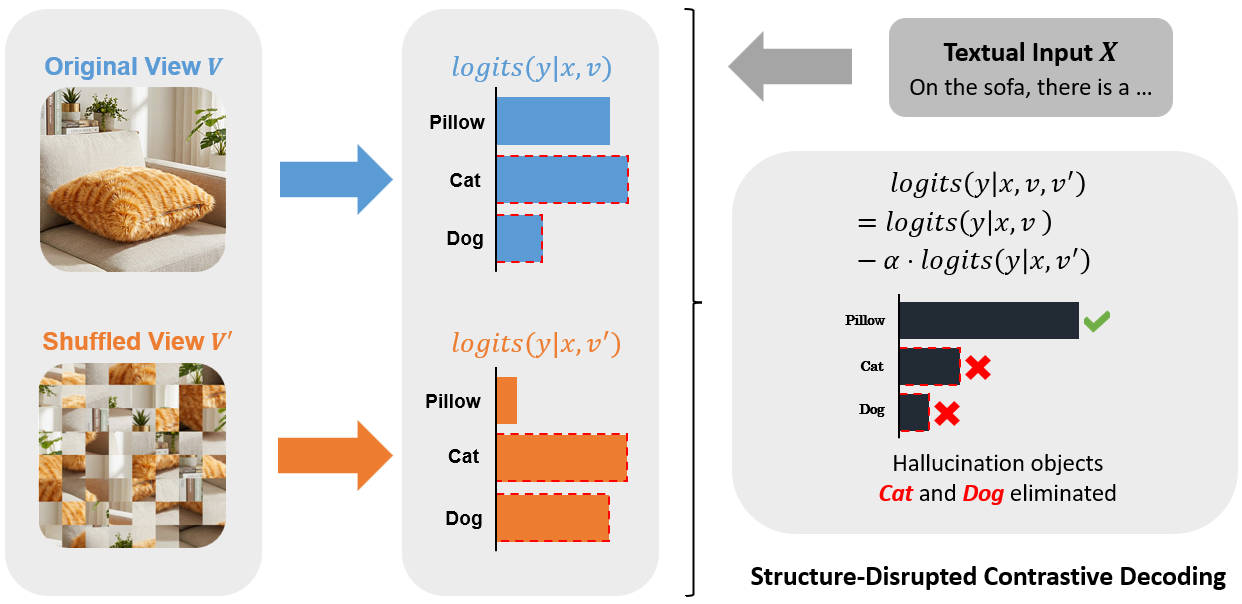}
    \caption{
        Overview of the proposed SDCD framework.
        Given a textual input $x$, the model performs contrastive decoding by jointly leveraging the original view $V$, a shuffled view $V'$.
        By suppressing texture-driven bias, SDCD effectively eliminates hallucinated objects during generation.
    }
    \label{fig:sdcd_overview}
\end{figure}

To verify the existence of this visual statistical bias, we conduct pilot studies. As illustrated in the example in Figure~\ref{fig:sdcd_overview}, a pillow with cat-fur-like textures induces the model to assign excessive confidence to hallucinated tokens in the decoding probability distribution, directly triggering the generation of non-existent objects. Notably, even when the global geometric structure of the image is completely disrupted through a shuffle operation, as long as local texture statistics are preserved, the model still maintains or even increases its confidence in hallucinated objects. In contrast, the generation behavior of real object tokens is highly sensitive to spatial perturbations, with their logit scores rapidly decreasing as geometric integrity is destroyed. Quantitative experiments show that real objects and hallucinations exhibit fundamentally different generation behaviors under structural disruptions. This divergence in structural sensitivity, termed \textit{Structure Sensitivity Divergence}, reveals a fundamental discrepancy in how the model relies on structural evidence when generating different types of tokens.

Building on this insight, we propose Structure-Disrupted Contrastive Decoding (SDCD), a training-free inference-time calibration strategy. As illustrated in Figure~\ref{fig:sdcd_overview}, the core of SDCD lies in leveraging a structure-disrupted view as a negative constraint. The shuffled view characterizes and suppresses misleading texture cues that induce hallucinations through structural disruption while the original view is retained as a baseline. By eliminating the influence of such biases, the model achieves more robust structural anchoring, effectively mitigating object hallucinations induced by visual statistical bias.

The main contributions of this paper are summarized as follows:

\begin{itemize}
    \item We identify that object hallucination in LVLMs is closely associated with a Bag-of-Patches driven visual bias, where Vision Encoders rely on local texture evidence in the absence of reliable global structural cues.

    \item We discover a Structure Sensitivity Divergence between real and hallucinated object tokens, and exploit this property to develop SDCD, a structure-disrupted contrastive decoding strategy that calibrates model's output by penalizing texture-driven biases.

    \item Experiments demonstrate that SDCD significantly mitigates hallucinations across multiple benchmarks and enhances the overall multimodal capabilities of LVLMs. These results validate its effectiveness and generality across different model architectures.
\end{itemize}

\section{Related Work}

\subsection{Object Hallucination and Mitigation in LVLMs}

Despite the significant strides Large Vision-Language Models (LVLMs) have made in multimodal understanding and reasoning \cite{liu2023visualinstructiontuning, zhu2023minigpt4enhancingvisionlanguageunderstanding, bai2023qwenvlversatilevisionlanguagemodel, wang2024qwen2vlenhancingvisionlanguagemodels, bai2025qwen25vltechnicalreport}, \textit{Object Hallucination}, where LVLMs generate object tokens inconsistent with the image content, continues to limit their reliability in real-world scenarios \cite{li2023evaluatingobjecthallucinationlarge, rohrbach2019objecthallucinationimagecaptioning, biten2021letclockbeachreducing}. The primary causes of such hallucinations are the models' language priors and statistical biases \cite{tong2024eyeswideshutexploring, guan2024hallusionbenchadvanceddiagnosticsuite, zhou2024analyzingmitigatingobjecthallucination, chen2024halcobjecthallucinationreduction}. While some approaches seek to reduce object hallucination through retraining on constructed datasets \cite{liu2024mitigatinghallucinationlargemultimodal, gunjal2024detectingpreventinghallucinationslarge}, a significant number of training-free methods have emerged. Some mitigate hallucinations by constructing various positive and negative views of inference images or instructions for contrastive decoding \cite{leng2023mitigatingobjecthallucinationslarge, wang2024mitigatinghallucinationslargevisionlanguage, favero2024multimodalhallucinationcontrolvisual, tong2025mitigatinghallucinationmultimodalllms}; others enhance image attention or modify abnormal attentional map \cite{liu2024payingattentionimagetrainingfree, huang2024operaalleviatinghallucinationmultimodal, zhu2024ibdalleviatinghallucinationslarge}; and some employ external models for object detection \cite{Yin_2024, zhou2024analyzingmitigatingobjecthallucination} or utilize stable diffusion models to generate views for contrastive decoding \cite{zhang2025selfcorrectingdecodinggenerativefeedback, cao2025cofidechallucinationresistantdecodingcoarsetofine}.

While current decoding-based interventions effectively prioritize the mitigation of language priors or the detection of attentional anomalies, they may still overlook misleading cues that may be inherent in the visual representations themselves (e.g., spurious texture statistics \cite{geirhos2022imagenettrainedcnnsbiasedtexture}). Consequently, they fail to address the root causes of visual encoding issues in image understanding.

\subsection{Bag-of-Patches Behavior from Vision Encoders to LVLMs}

Modern LVLMs typically employ Vision Transformers and their variants \cite{radford2021learningtransferablevisualmodels, zhai2023sigmoidlosslanguageimage, sun2023evaclipimprovedtrainingtechniques} as Vision Encoders. These models divide an image into a sequence of patches and aggregate features via self-attention mechanisms. Despite their superior performance in semantic classification tasks, prior studies have demonstrated that their visual representations inherently exhibit a distinct Bag-of-Patches behavior, showing high insensitivity to the spatial arrangement of patches \cite{naseer2021intriguingpropertiesvisiontransformers, geirhos2022imagenettrainedcnnsbiasedtexture}. Even after significant shuffling of the image structure, these models retain high recognition performance, indicating that their decision-making process relies more on \textit{local texture statistics} rather than explicit modeling of the object's global geometry or spatial relationships \cite{yuksekgonul2023visionlanguagemodelsbehavelike}.

This lack of structural perception is further amplified in LVLMs. Recent research has found that when the global structure of an input image is disrupted, LVLMs often fail to detect the anomaly and continue to generate plausible descriptions that are inconsistent with the actual visual content \cite{tong2024eyeswideshutexploring, Parcalabescu_2022}. Building upon these insights, this paper proposes Structure-Disrupted Contrastive Decoding (SDCD). By constructing \textit{semantically preserved but structurally disrupted views}, SDCD explicitly amplifies the difference in model response to structural absence, thereby identifying and suppressing texture-driven hallucination generation.

\section{Methods}

\subsection{Vision Encoder's Bag-of-Patches Behavior}

Although Transformer-based Vision Encoders such as ViT \cite{dosovitskiy2020imageViT} and CLIP \cite{radford2021learningtransferablevisualmodels} incorporate positional encodings to retain spatial information, they often exhibit an inherent "Bag-of-Patches" behavior\cite{qin2023understandingimprovingrobustnessvision,gu2022visiontransformersrobustpatch}. Specifically, these models tend to aggregate features based on local texture co-occurrences rather than global geometric structures or compositional relationships. This leads to a critical perceptual asymmetry: while the encoder is highly robust to local semantic content, it fails to perceive the spatial arrangement of those patches. We hypothesize that this behavior is a primary driver of hallucinations, as LVLMs may generate descriptions based on isolated visual cues even when the global structure is non-existent or contradictory.

To explicitly decouple visual semantics from structural information, we leverage the aforementioned Bag-of-Patches behavior to construct a structure-disrupted view. Specifically, given an input image $I \in \mathbb{R}^{H \times W \times C}$ and shuffle patch size $S$, we reshape $I$ into a sequence of $N = HW/S^2$ patches $V = \{v_1, v_2, \dots, v_N\}$, where each $v_i$ represents a local region of the image. We introduce a random permutation function $\pi(\cdot)$ to shuffle the patch indices in $v$, thereby generating the shuffled view $V'$:

\begin{equation}
    V' = \{v_{\pi(1)}, v_{\pi(2)}, \dots, v_{\pi(N)}\},
\end{equation}

where $\pi$ is a bijection from $\{1, \dots, N\}$ to $\{1, \dots, N\}$. This operation mathematically \textit{severs} the geometrical dependencies and spatial correlations between adjacent patches, yet fully \textit{preserves} all local texture features and object semantics in a set-theoretic sense. Consequently, $V'$ serves as a perfect control variable: it contains all the \textit{patch-level evidence }required for generation but eliminates the grammatical structure needed to correctly compose these ingredients.

\begin{figure}[t]
    \centering
    \begin{subfigure}[t]{0.31\linewidth}
        \centering
        \includegraphics[width=\linewidth]{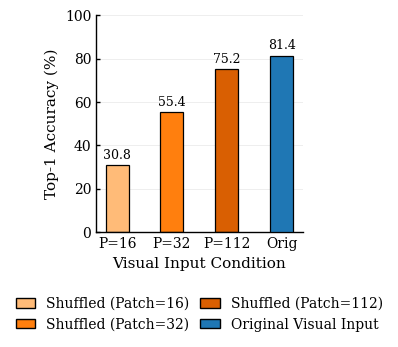}
        \caption{ViT}
        \label{fig:vit_bop}
    \end{subfigure}
    \hfill
    \begin{subfigure}[t]{0.66\linewidth}
        \centering
        \includegraphics[width=\linewidth]{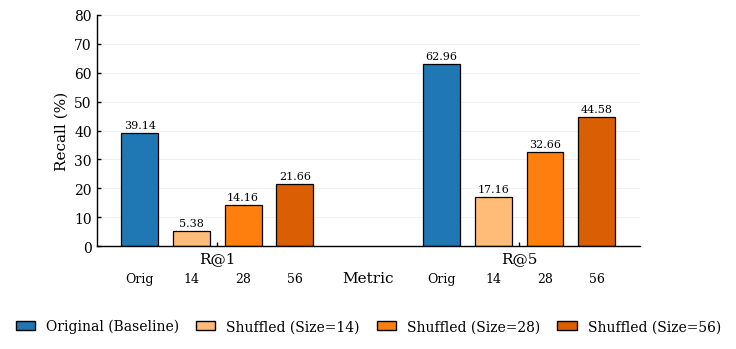}
        \caption{CLIP}
        \label{fig:clip_bop}
    \end{subfigure}
    
    \caption{
        Quantitative analysis of the Bag-of-Patches behavior. 
        (a) Top-1 Accuracy of ViT-B/16 on a random subset of 500 images from the ImageNet validation set under varying patch shuffling granularities.
        (b) Image-Text Retrieval performance (R@1 and R@5) of CLIP-ViT-L/14. We extract visual features from original and shuffled views and compute cosine similarity against text features from 5 corresponding captions per image.
        Despite severe structural disruption, both models retain strong semantic performance.
    }
    \label{fig:bop_analysis}
\end{figure}

To quantitatively examine the Bag-of-Patches hypothesis, we analyze the robustness of standard ViT-B/16 and CLIP-ViT-L/14 under controlled structural disruption. Specifically, we construct an original view $V$ and a shuffled view $V'$ by shuffling image patches, and evaluate ViT on ImageNet classification and CLIP on image–text retrieval using the corresponding visual representations. As shown in Figure~\ref{fig:bop_analysis}, even when global geometric structure is severely corrupted, both models retain surprisingly strong performance instead of collapsing. In particular, ViT preserves substantial classification accuracy across a wide range of shuffle scales, while CLIP maintains non-trivial retrieval capability despite the loss of spatial coherence. These results reveal a pronounced texture robustness: rather than relying on holistic geometric structure, modern Vision Encoders can sustain recognition by aggregating local texture statistics. This behavior suggests that existing ViT-style encoders have largely degenerated into \textit{visual bag-of-words} models, where semantics are assembled from local cues with limited structural grounding.

\begin{figure}[t]
    \centering
    \begin{subfigure}[t]{0.48\linewidth}
        \centering
        \includegraphics[width=\linewidth]{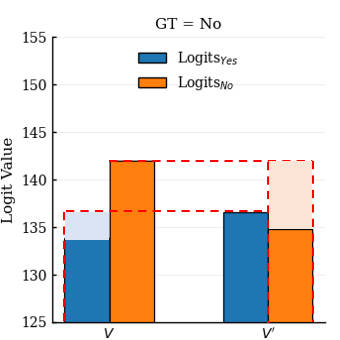}
        \caption{GT = No}
        \label{fig:llava_gt_no}
    \end{subfigure}
    \hfill 
    \begin{subfigure}[t]{0.48\linewidth}
        \centering
        \includegraphics[width=\linewidth]{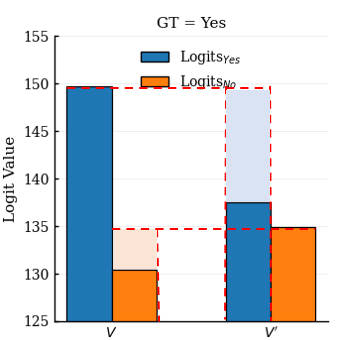}
        \caption{GT = Yes}
        \label{fig:llava_gt_yes}
    \end{subfigure}
    
    \caption{
    An illustration of Structure Sensitivity Divergence in LVLM decoding.
    We analyze the logit dynamics of Yes and No tokens under the original view ($V$) and the shuffled view ($V'$).
    (a) For ground-truth No cases (potential hallucinations), the confidence of the incorrect Yes token often increases under $V'$, indicating a texture-dominated response when global structure is removed.
    (b) For ground-truth Yes cases (real objects), the confidence of the correct Yes token drops sharply under $V'$, revealing a strong dependence on global geometric structure (structural penalty).
    The shaded regions highlight the difference in token logits between the original view ($V$) and the shuffled view ($V'$).
    This asymmetric response characterizes the Structure Sensitivity Divergence between real and hallucinated objects.
}

    \label{fig:llava_logits_analysis}
\end{figure}

\subsection{Structure Sensitivity Divergence in LVLMs}
This Bag-of-Patches behavior inevitably permeates into the generative space of LVLMs. 
To investigate its impact, we conducted a comparative study on 1,000 images. 
We queried the model using the prompt ``\textit{Is there a <object> in the image?}'' regarding both ground-truth existent and non-existent objects under the original view $V$ and the shuffled view $V'$. During this process, we monitor the logit dynamics of Yes/No tokens. 
As shown in Figure~\ref{fig:llava_logits_analysis}, we observed a significant \textit{Structure Sensitivity Divergence}. 
For ground-truth existent objects, structural disruption introduces a distinct \textit{Structural Penalty}. 
Specifically, the logit for the correct answer drops significantly and rapidly approaches that of the incorrect answer, indicating that the recognition of real objects relies heavily on cross-patch structural consistency. 
In contrast, for potential hallucinated objects, structural disruption triggers a \textit{Texture Unleashed} phenomenon. 
Here, local textures dominate decision-making and significantly elevate the Yes confidence, which on average even surpasses No. 
This asymmetric sensitivity, in which real objects degenerate due to structural loss while hallucinated objects are enhanced, reveals the intrinsic structural dependency bias of LVLMs and constitutes the core motivation for SDCD to utilize the visually distorted view for hallucination suppression.

\subsection{Structure-Disrupted Contrastive Decoding (SDCD)}

Building upon the analysis of Structure Sensitivity Divergence, we identify that the shuffled view $V'$ undermines global structural constraints. This degradation induces the model to rely on local textures for generation, which is a behavior intrinsically linked to object hallucination. To address this, we propose Structure-Disrupted Contrastive Decoding (SDCD). 
SDCD leverages shuffled view $V'$ to eliminate texture biases stemming from the Bag-of-Patches behavior within the generated logits. Specifically, given a textual query $x$ and the original view $V$, the model yields two independent output distributions: $p_{\theta}(y|V, x)$ conditioned on the original view, and $p_{\theta}(y|V', x)$ conditioned on the structure-disrupted (i.e., patch-shuffled) view. Since $V'$ disrupts global geometric constraints, the resulting distribution $p_{\theta}(y|V', x)$ essentially characterizes the model's hallucination-prone distribution derived solely from local textural cues.

To suppress this bias, SDCD performs contrastive calibration on the two decoding results within the logit space. The final decoding distribution is defined as:

\begin{equation}
\small 
\begin{aligned}
p_{\text{SDCD}}(y_t \mid V, V', x)
&= \operatorname{softmax}\big(
(1+\alpha)\,\operatorname{logit}_{\theta}(y_t \mid V, x) \\
&\qquad {}- \alpha\,\operatorname{logit}_{\theta}(y_t \mid V', x)
\big),
\end{aligned}
\end{equation}

where $\alpha \geq 0$ is a contrastive hyperparameter controlling the penalty strength on texture bias. SDCD serves as a Structure-Aware Corrective Mechanism:
\begin{itemize}
    \item For ground-truth objects, the generation logits of their corresponding tokens are highly sensitive to structural disruption and drop significantly under $V'$. In this case, contrastive calibration primarily preserves and reinforces the structural consistency evidence from the original view.
    \item For potential hallucinated objects, the corresponding tokens may maintain high logits or even exhibit an increase due to texture induction after structural disruption. In such scenarios, the contrastive term directly suppresses their probability in the final decoding distribution.
\end{itemize}
By penalizing tokens that retain high confidence under structure-less conditions, SDCD effectively mitigates the interference of texture-dominated biases in the generation process.

\section{Experiment}

\subsection{Experimental Settings}

\subsubsection{Datasets \& Evaluation Metrics}

To comprehensively evaluate the effectiveness of our method, we adopt a three-fold evaluation protocol. 
We first utilize POPE \cite{li2023evaluatingobjecthallucinationlarge} for strictly discriminative hallucination assessment. 
Subsequently, to ensure that the mitigation of hallucinations does not come at the expense of general multimodal capabilities, we incorporate the MME benchmark \cite{fu2025mmecomprehensiveevaluationbenchmark}. 
Finally, we employ CHAIR \cite{rohrbach2019objecthallucinationimagecaptioning} to examine object hallucinations in open-ended generation tasks.

\paragraph{POPE}
We adopt POPE as the core benchmark for object hallucination
evaluation.
POPE formulates hallucination detection as a balanced binary
object-existence probing task by ask LVLMs questions such as "\textit{Is there a <object> in the image?}", which enables controlled and quantitative
assessment.
Following prior work, we report results under its three sampling
settings, namely Random, Popular, and Adversarial, across three
datasets, including MSCOCO \cite{lin2015microsoftcococommonobjects}, A-OKVQA \cite{schwenk2022aokvqabenchmarkvisualquestion}, and GQA \cite{hudson2019gqanewdatasetrealworld}.
This design allows us to systematically examine hallucinations induced
by random negative sampling, frequent-object bias, and object
co-occurrence bias, respectively.

\paragraph{MME}
To verify that hallucination mitigation does not compromise overall
multimodal capabilities, we further evaluate our method on the
MME benchmark.
We use the Hallucination subset, which consists of Existence, Count,
Position, and Color tasks, to assess object-level and attribute-level
hallucinations.
In addition, we report results on the full Perception and Cognition
subsets to examine potential effects on general reasoning and
perception performance.

\paragraph{CHAIR}
To evaluate hallucinations in open-ended generation, we employ the CHAIR metric~\cite{rohrbach2019objecthallucinationimagecaptioning}, which identifies hallucinations by matching generated objects against ground-truth labels.
We compute both instance-level ($\text{CHAIR}_I$) and sentence-level ($\text{CHAIR}_S$) metrics, formulated as:
\begin{equation}
    \text{CHAIR}_I = \frac{|\{\text{hallucinated objects}\}|}{|\{\text{all mentioned objects}\}|}, 
\end{equation}
\begin{equation}
    \text{CHAIR}_S = \frac{|\{\text{hallucinated captions}\}|}{|\{\text{all captions}\}|}.
\end{equation}
We conduct experiments on 500 randomly sampled instances from the MSCOCO validation set, using the prompt ``\textit{Please help me describe the image in detail.}'' with a maximum generation length of 512 tokens.

\subsubsection{Method Baselines}
We compare SDCD with three distinct categories of decoding strategies: 
\begin{itemize}
    \item \textbf{Regular Decoding:} serving as the fundamental baseline, utilizing the vanilla LVLMs for autoregressive generation without any external intervention or distribution adjustment. It reflects the model's intrinsic performance and its susceptibility to inherent biases.
    \item \textbf{VCD} \cite{leng2023mitigatingobjecthallucinationslarge}: serving as the primary internal baseline, this method represents calibration techniques based on visual uncertainty. It introduces a visually distorted view by superimposing Gaussian noise onto the original image, subsequently employing contrastive decoding between the original and distorted inputs to calibrate the output logits and mitigate statistical biases.
    \item \textbf{PAI} \cite{liu2024payingattentionimagetrainingfree}: representing prior-based inference interventions. It enhances the model's focus by amplifying the attention weights allocated to image tokens. Concurrently, it constructs a negative view derived from a pure-textual context to perform contrastive decoding, thereby effectively suppressing language priors and forcing the model to ground its response in visual evidence.

\end{itemize}
Through comparison with these three categories, we aim to demonstrate the comprehensive advantages of SDCD in terms of both efficiency and effectiveness.

\subsubsection{LVLM Baselines}
To verify the applicability of SDCD across different projector architectures, we select three representative 7B models. LLaVA-1.5 \cite{liu2023visualinstructiontuning} and Qwen2.5-VL-7B-Instruct \cite{bai2025qwen25vltechnicalreport} represent MLP architectures that preserve spatial structure, while Qwen-VL \cite{bai2023qwenvlversatilevisionlanguagemodel} represents a compression architecture based on a Resampler. This model selection strategy allows us to deeply investigate the manifestation differences of the Vision Encoder's Bag-of-Patches behavior across architectures and the targeted improvements of SDCD.

\subsubsection{Implementation Details}
All experiments are conducted on a single NVIDIA A100 GPU in a training-free manner. 
To construct the structure-disrupted view $V'$, we employ a shuffle grid size\footnote{The results of the ablation study on different shuffle sizes $S$ are detailed in Section~\ref{sec:ablation_size}.} $S$ of $14 \times 14$ for LLaVA-1.5 and Qwen-VL. For Qwen2.5-VL, we adjust the granularity to $28 \times 28$ to align with the token pooling mechanism in its projector architecture. 
Regarding decoding configurations, we set the contrastive hyperparameter\footnote{The results of the ablation study on different $\alpha$ are detailed in Section~\ref{sec:ablation_alpha}.} $\alpha = 2.0$ and the Adaptive Plausibility Constraint threshold $\beta = 0.1$. To further enhance visual focus, the attention weights allocated to image tokens are increased by 0.6 across both views. 
Generation employs nucleus sampling with a temperature of 1.0 and top-$p$ of 0.9.

\subsection{Main Results}
\subsubsection{Results on POPE}
We categorize the experimental results on the POPE benchmark into two groups based on the projector architectures of the LVLMs. The first group comprises models utilizing MLP projectors, such as LLaVA and Qwen2.5-VL, representing the most prevalent architecture currently. The second group consists of Qwen-VL, which employs a Resampler as its projector.

\paragraph{Results on LVLMs with MLP Projectors}

\begin{table*}[t]
  \centering
  \small
  \setlength{\tabcolsep}{3pt}
  \renewcommand{\arraystretch}{1.05}
  \resizebox{\textwidth}{!}{
  \begin{tabular}{lll|cccc|cccc|cccc}
    \hline
    \multirow{2}{*}{\textbf{Dataset}}
    & \multirow{2}{*}{\textbf{Model}}
    & \multirow{2}{*}{\textbf{Decoding}}
    & \multicolumn{4}{c|}{\textbf{Random}}
    & \multicolumn{4}{c|}{\textbf{Popular}}
    & \multicolumn{4}{c}{\textbf{Adversarial}} \\
    \cline{4-15}
     & & 
     & Acc. & Prec. & Rec. & F1
     & Acc. & Prec. & Rec. & F1
     & Acc. & Prec. & Rec. & F1 \\
    \hline

\multirow{6}{*}{MSCOCO}
& \multirow{3}{*}{LLaVA1.5}
& Regular
& 82.93 & 92.01 & 72.13 & 80.87
& 81.10 & 87.90 & 72.13 & 79.24
& 78.63 & 82.96 & 72.07 & 77.13 \\
& & VCD
& 84.87 & 92.52 & 75.87 & 83.37
& 82.43 & 87.34 & 75.87 & 81.20
& 79.90 & 82.52 & 75.87 & 79.06 \\
& & SDCD
& \textbf{85.90} & \textbf{93.46} & \textbf{77.20} & \textbf{84.56}
& \textbf{83.43} & \textbf{88.19} & \textbf{77.20} & \textbf{82.33}
& \textbf{80.87} & \textbf{83.36} & \textbf{77.13} & \textbf{80.12} \\
\cline{2-15}

& \multirow{3}{*}{Qwen2.5-VL}
& Regular
& 81.73 & 98.37 & 64.53 & 77.94
& 81.23 & \textbf{97.66} & 64.00 & 77.33
& 80.83 & 95.84 & 64.47 & 77.08 \\
& & VCD
& 82.67 & 98.80 & 66.13 & 79.23
& 82.97 & 97.50 & 67.67 & 79.89
& 81.77 & \textbf{96.40} & 66.00 & 78.35 \\
& & SDCD
& \textbf{85.63} & \textbf{98.90} & \textbf{72.07} & \textbf{83.38}
& \textbf{85.07} & 97.05 & \textbf{72.33} & \textbf{82.89}
& \textbf{83.93} & 94.57 & \textbf{72.00} & \textbf{81.76} \\
\hline

\multirow{3}{*}{A-OKVQA}
& \multirow{3}{*}{LLaVA1.5}
& Regular
& 84.03 & 87.68 & 79.20 & 83.22
& 80.23 & \textbf{80.87} & 79.20 & 80.03
& 74.27 & \textbf{72.33} & 78.60 & 75.34 \\
& & VCD
& 85.03 & 87.09 & 82.27 & 84.61
& 80.60 & 79.61 & 82.27 & 80.92
& \textbf{74.90} & 71.75 & 82.13 & \textbf{76.59} \\
& & SDCD
& \textbf{86.30} & \textbf{87.79} & \textbf{84.33} & \textbf{86.03}
& \textbf{80.93} & 78.96 & \textbf{84.33} & \textbf{81.56}
& 73.93 & 70.19 & \textbf{83.20} & 76.14 \\
\hline

\multirow{3}{*}{GQA}
& \multirow{3}{*}{LLaVA1.5}
& Regular
& 83.60 & 87.11 & 78.87 & 82.79
& 77.87 & \textbf{77.32} & 78.87 & 78.09
& 75.17 & \textbf{73.32} & 79.13 & 76.11 \\
& & VCD
& 84.83 & 86.56 & 82.47 & 84.47
& \textbf{78.07} & 75.80 & 82.47 & \textbf{78.99}
& \textbf{75.50} & 72.49 & 82.20 & \textbf{77.04} \\
& & SDCD
& \textbf{85.70} & \textbf{87.16} & \textbf{83.73} & \textbf{85.41}
& 76.47 & 73.11 & \textbf{83.73} & 78.06
& 74.50 & 70.50 & \textbf{84.27} & 76.77 \\
\hline
  \end{tabular}
  }
  \caption{Results for LLaVA1.5 and Qwen2.5-VL on POPE (values in \%).}
  \label{tab:pope_llava_qwen25}
\end{table*}

Table \ref{tab:pope_llava_qwen25} presents the experimental results on the POPE benchmark for representative models employing MLP projectors (LLaVA-1.5 and Qwen2.5-VL). Overall, our proposed SDCD outperforms both standard decoding (Regular) and the visual contrastive decoding baseline (VCD) across the majority of settings.

For LLaVA-1.5, SDCD achieves the best Accuracy and F1 scores across all three evaluation dimensions on MSCOCO. Specifically, under the Random setting, SDCD reaches an accuracy of 85.90\%, surpassing Regular (82.93\%) and VCD (84.87\%). On A-OKVQA and GQA, SDCD maintains a similar advantage, particularly in Recall metrics. Comparable improvements are observed with Qwen2.5-VL. This model exhibits a distinct ``high precision, low recall'' characteristic in Regular mode (e.g., 98.37\% Precision versus 64.53\% Recall in MSCOCO Random), reflecting an extreme conservative tendency. SDCD increases Recall by approximately 7.5 percentage points to 72.07\% while maintaining high precision, thereby achieving the highest F1 score (83.38\%). This indicates that SDCD effectively corrects the model's bias towards over-refusal, mitigating missed detections without introducing additional hallucinations.

\paragraph{Results on LVLMs with Resampler-based Projectors}

\begin{table*}[t]
  \centering
  \small
  \setlength{\tabcolsep}{3pt}
  \renewcommand{\arraystretch}{1.05}
  \resizebox{\textwidth}{!}{
  \begin{tabular}{lll|cccc|cccc|cccc}
    \hline
    \multirow{2}{*}{\textbf{Dataset}}
    & \multirow{2}{*}{\textbf{Model}}
    & \multirow{2}{*}{\textbf{Decoding}}
    & \multicolumn{4}{c|}{\textbf{Random}}
    & \multicolumn{4}{c|}{\textbf{Popular}}
    & \multicolumn{4}{c}{\textbf{Adversarial}} \\
    \cline{4-15}
     & & 
     & Acc. & Prec. & Rec. & F1
     & Acc. & Prec. & Rec. & F1
     & Acc. & Prec. & Rec. & F1 \\
    \hline

\multirow{3}{*}{MSCOCO}
& \multirow{3}{*}{Qwen-VL}
& Regular
& 84.43 & 95.27 & 72.47 & 82.32
& 83.87 & 94.10 & 72.27 & 81.75
& 82.40 & 89.84 & 73.07 & 80.59 \\
& & VCD
& \textbf{86.47} & \textbf{95.87} & \textbf{75.87} & \textbf{84.70}
& \textbf{85.80} & 94.37 & \textbf{75.93} & \textbf{84.15}
& \textbf{83.90} & \textbf{89.95} & \textbf{76.33} & \textbf{82.58} \\
& & SDCD
& 85.17 & 95.67 & 73.67 & 83.24
& 84.93 & \textbf{94.79} & 73.93 & 83.07
& 82.67 & 89.26 & 74.27 & 81.08 \\
\hline

\multirow{3}{*}{A-OKVQA}
& \multirow{3}{*}{Qwen-VL}
& Regular
& 60.87 & 62.79 & 53.33 & 57.68
& 60.40 & \textbf{62.06} & 53.53 & 57.48
& 79.27 & \textbf{79.99} & 78.07 & 79.01 \\
& & VCD
& \textbf{61.83} & 63.13 & 55.60 & 59.13
& \textbf{61.10} & 61.90 & \textbf{56.93} & \textbf{59.22}
& \textbf{80.13} & 79.35 & 81.47 & \textbf{80.39} \\
& & SDCD
& 61.80 & \textbf{63.27} & \textbf{56.27} & \textbf{59.56}
& 60.13 & 60.86 & 56.80 & 58.76
& 79.63 & 78.29 & \textbf{82.00} & 80.10 \\
\hline

\multirow{3}{*}{GQA}
& \multirow{3}{*}{Qwen-VL}
& Regular
& 58.93 & 61.13 & 49.07 & 54.44
& \textbf{58.87} & \textbf{59.87} & 53.80 & 56.67
& 75.47 & 78.13 & 70.73 & 74.25 \\
& & VCD
& \textbf{59.70} & \textbf{61.41} & \textbf{52.20} & \textbf{56.43}
& 58.83 & 58.65 & 56.73 & 57.68
& \textbf{77.30} & \textbf{78.98} & \textbf{74.40} & \textbf{76.62} \\
& & SDCD
& 58.97 & 60.70 & 50.87 & 55.35
& 57.47 & 57.22 & \textbf{59.20} & \textbf{58.19}
& 74.47 & 75.45 & 72.53 & 73.96 \\
\hline
  \end{tabular}
  }
  \caption{Results for Qwen-VL on POPE (values in \%).}
  \label{tab:pope_qwenvl}
\end{table*}

Table \ref{tab:pope_qwenvl} reports the performance of Qwen-VL, which utilizes a Resampler-based projector. Unlike the MLP-based models, Qwen-VL exhibits a slightly different trend where SDCD and VCD exhibit complementary strengths across different settings. While VCD holds a slight edge on MSCOCO, SDCD achieves the highest F1 scores in the more challenging settings of A-OKVQA (Random) and GQA (Popular). We attribute this to the architectural distinction: MLP projectors (as in LLaVA) preserve patch-wise independence, retaining local biases that SDCD effectively targets. In contrast, Qwen-VL's Cross-Attention aggregates visual information, which inherently mixes local textures and partially dampens the structural perturbation signals (see Section \ref{sec:arch_analysis} for detailed analysis).

Nevertheless, SDCD retains distinct advantages. In the Random setting of A-OKVQA, SDCD achieves the highest F1 score (59.56\%) and Precision (63.27\%). Similarly, on GQA (Popular setting), SDCD attains the best F1 score of 58.19\%. Furthermore, SDCD continues to excel in Recall improvement, reaching a peak Recall of 82.00\% in the A-OKVQA Adversarial setting. This suggests that even within Resampler-based projectors, SDCD remains a robust strategy for complex reasoning tasks and for mitigating missed detections.

\subsubsection{Result on MME}
\begin{table}[t]
  \centering
  \small
  \setlength{\tabcolsep}{2pt}
  \begin{tabular}{lcccccc}
    \hline
    \textbf{Method} & \textbf{Exist.} & \textbf{Count} & \textbf{Pos.} & \textbf{Color} & \textbf{Perc.} & \textbf{Cog.} \\
    \hline
    Regular        & 175.00          & \textbf{128.33} & 85.00           & \textbf{155.00} & 1229.93          & 307.14 \\
    VCD & 175.00          & 106.67          & 111.67          & 136.67          & 1292.01          & 286.43 \\
    SDCD    & \textbf{190.00} & 126.67          & \textbf{125.00} & 143.33          & \textbf{1348.35} & \textbf{338.93} \\
    \hline
  \end{tabular}
  \caption{Results on the MME benchmark. All experiments are running LLaVA-1.5. We report scores for specific hallucination-related subtasks (Existence, Count, Position, Color) alongside the aggregated Perception and Cognition scores. Best results are highlighted in bold.}
  \label{tab:mme_results}
\end{table}

On the MME benchmark\footnote{Full results of MME are detailed in Section~\ref{sec:detail_mme}.}, we observe that our SDCD method leads to consistent and substantial improvements across perception-related tasks. Show in Table \ref{tab:mme_results}, SDCD achieves the most pronounced gains in structure-sensitive categories such as existence, position, landmark, and celebrity recognition. Compared to regular decoding and Gaussian Noise-based VCD, SDCD achieves the highest overall perception score, indicating stronger visual grounding and structural awareness. Beyond perception, SDCD also demonstrates a notable advantage in cognition-oriented tasks. While VCD slightly degrades cognition performance, SDCD significantly improves overall cognition, with especially large gains in code reasoning. This suggests that reinforcing structural consistency through shuffle-based contrastive views not only mitigates hallucinations but also stabilizes higher-level multimodal reasoning.

\subsubsection{Result on CHAIR}

\begin{table}[t]
  \centering
  \small
  \setlength{\tabcolsep}{6pt}
  \renewcommand{\arraystretch}{1.05}
  \begin{tabular}{lccc}
    \hline
    \textbf{Method} & $\textbf{CHAIR}_S$ $\downarrow$ & $\textbf{CHAIR}_I$ $\downarrow$ & \textbf{F1} $\uparrow$ \\
    \hline
    Regular & 55.6 & 17.3 & 72.2 \\
    VCD     & 57.0 & 16.9 & \textbf{74.4} \\
    PAI     & 38.2 & 11.0 & 73.4 \\
    \textbf{SDCD}   & \textbf{18.6} & \textbf{6.4} & 70.7 \\
    \hline
  \end{tabular}
  \caption{Results on the CHAIR benchmark under nucleus sampling running LLaVA. Lower $\textbf{CHAIR}_S$ and $\textbf{CHAIR}_I$ indicate fewer hallucinated sentences and objects, while higher F1 reflects a better balance between precision and recall.}
  \label{tab:chair_main}
\end{table}

We evaluate the performance of Regular, VCD, PAI, and SDCD on the CHAIR benchmark under a unified nucleus sampling decoding setting, as detailed in Table \ref{tab:chair_main}. As observed, SDCD achieves a significant advantage in both $\textbf{CHAIR}_S$ and $\textbf{CHAIR}_I$ metrics. Compared to Regular, VCD, and PAI, SDCD yields substantial reductions in these scores, indicating its effectiveness in suppressing hallucinated objects during the generation process. Notably, the $\text{CHAIR}_S$ and $\text{CHAIR}_I$ score of SDCD decrease to 18.6 and 6.4, demonstrating a robust capability in inhibiting hallucinated instances. These results confirm that SDCD significantly mitigates the hallucination rate through structural constraints.

\subsection{Discussion: Impact of Visual Projectors}
\label{sec:arch_analysis}

We observe a distinct architectural discrepancy on the POPE benchmark: the performance gains achieved by SDCD on LLaVA-1.5 and Qwen2.5-VL are notably superior to those on Qwen-VL. This phenomenon suggests that there is a strong correlation between the design of \textit{projector} and the \textit{effectiveness of hallucination suppression}, warranting further analysis at the architectural level.

Both LLaVA-1.5 and Qwen2.5-VL employ MLPs as the projector,
which performs a patch-wise, one-to-one mapping from Vision Encoder features to visual tokens,
without explicitly encoding spatial relationships across patches.
As a result, visual information is introduced into the language model mainly through local features.
Without additional global structural constraints, this projection scheme is more likely
to inherit the Bag-of-Patches behavior of Vision Encoders, encouraging reliance on local
texture statistics during generation.
By disrupting patch-level spatial structure through Patch Shuffling, SDCD attenuates this
local-statistics-driven bias, leading to more pronounced improvements in models that adopt
MLP-based projectors.

In contrast, Qwen-VL adopts a \textbf{cross-attention} mechanism for cross-modal fusion,
in which a set of intermediate visual tokens aggregates information from patch-level
visual features.
This design mixes information across patches before the visual representations are
passed to the language model, rather than preserving independent patch-wise tokens.
As a result, the model exhibits weaker reliance on the local textures of individual patches.
Under this architecture, the structural disruption introduced by Patch Shuffling overlaps with the intrinsic fusion process, resulting in gains from SDCD that do not consistently surpass those of VCD across all settings.

From the perspective of architectural evolution, as LVLMs advance towards higher input resolutions and finer-grained visual modeling capabilities, the design of patch-wise projection, in which visual features are directly fed into the language model, has been widely adopted across numerous models. In this context, the dominance of local statistics together with insufficient global structural constraints is likely to become more pronounced. The consistent improvements achieved by SDCD on Qwen2.5-VL demonstrate its effective adaptability to this class of projection methods, providing a viable inference-time solution for mitigating the resulting visual biases.

\section{Conclusion}
In this work, we investigated object hallucination in LVLMs from the perspective of the texture bias and identified that the Bag-of-Patches behavior of Vision Encoders induces a texture-dominated, structure-insensitive generation behavior. Based on the observed Structure Sensitivity Divergence between real and hallucinated objects tokens under structural disruption, we proposed SDCD, a training-free contrastive decoding strategy that suppresses texture-driven hallucinations by penalizing tokens insensitive to structural corruption. Extensive experiments across POPE, MME, and CHAIR benchmarks demonstrate that SDCD effectively reduces hallucinations while preserving or even improving overall multimodal reasoning performance. We hope this study sheds light on the role of visual structure in LVLM decoding and inspires future work on structure-aware inference mechanisms.

\section*{Limitations}

SDCD may exhibit different levels of effectiveness across projector architectures. When visual information is fed into the language model in a relatively independent patch-level form, such as in LLaVA where an MLP performs a patch-wise, one-to-one projection from visual patches to tokens, SDCD tends to achieve more pronounced gains. In contrast, for models that employ Resampler-based projectors, visual features are mixed across patches before entering the language model, which may partially attenuate the contrastive signal introduced by structural disruption. As a result, the advantages of SDCD can vary across model architectures, and its effectiveness is more likely to be fully realized in models that preserve patch-level visual representations.

\section*{Acknowledgments}

We thank our colleagues in the Department of Electrical and Computer Engineering at the University of California, Santa Barbara and the open-source community for helpful discussions and resources.

\bibliography{custom}
\newpage
\appendix
\section{Detailed Experimental Settings}

In the implementation of SDCD, for LLaVA-1.5 and Qwen-VL, we construct the shuffled view $V'$ using a shuffle grid size of $14 \times 14$ to maximally disrupt global structure while preserving local texture semantics. For Qwen2.5-VL, due to the image token pooling mechanism in its visual encoding stage, we correspondingly set the patch granularity to $28 \times 28$ to match its projector Structure. The contrastive hyperparameter $\alpha$ is set to $2.0$. Furthermore, to prevent over-calibration from compromising sentence fluency, we adopt the Adaptive Plausibility Constraint, setting $\beta = 0.1$ to filter out low-probability candidate tokens. We choose nucleus sampling with Top-p 0.9 and temperature 1.0. To enhance visual focus during decoding, we increased the attention weights of all tokens towards image tokens by 0.6 in both views. All experiments are conducted on a single NVIDIA A100 GPU, requiring no additional model training or external auxiliary models.

\section{Ablation Study}
\subsection{Effect of Shuffle Grid Size $S$}
\label{sec:ablation_size}

To investigate the optimal granularity for structural disruption, we examine the impact of varying Shuffle Grid Sizes (denoted as pixel dimensions $S \times S$ of the shuffling unit) on model performance. As shown in Table~\ref{tab:shuffle_size_ablation}. experimental results indicate that for the LLaVA-1.5 model, optimal performance is achieved when the shuffling granularity is set to $S=14$. As the granularity increases to $S=28$ and $S=56$, the model exhibits a consistent trend of performance degradation: the F1 score drops by approximately 3\% and 4\%, respectively, while the decline in Recall further widens to approximately 4\% and 7\%. This phenomenon suggests that as the shuffling granularity becomes coarser, the structural distinguishability between $V'$ and $V$ diminishes. This reduction weakens the effective constraints provided by SDCD during the contrastive decoding process, ultimately manifesting as a systematic decline in model performance.

Current mainstream LVLMs (e.g., the LLaVA series) typically employ CLIP-ViT-L/14 as the Vision Encoder, which has a native Patch Size of $14 \times 14$. When the Shuffle Grid Size is set to $S=14$, the shuffling operation targets the minimal semantic units of the visual features. This configuration maximally disrupts cross-token spatial adjacency relationships, causing the generated shuffled view $V'$ to lose its global geometric structure while preserving the local texture semantics within each individual patch. In contrast, when the Grid Size increases to $S=28$ or $S=56$, shuffling is performed on larger blocks composed of multiple adjacent patches. In these cases, the local spatial structure within these blocks is preserved holistically, resulting in $V'$ retaining a strong structural consistency with $V$. This structural residue directly reduces the intensity of the structural contrast between $V$ and $V'$, thereby hampering SDCD's ability to fully suppress hallucinations arising from structural biases.

\begin{table}[t]
  \centering
  \small
  \setlength{\tabcolsep}{4pt}
  \begin{tabular}{ccccc}
    \hline
    \textbf{Shuffle Size} & \textbf{Precision} & \textbf{Recall} & \textbf{F1} & \textbf{Accuracy} \\
    \hline
    14 & \textbf{93.46} & \textbf{77.20} & \textbf{84.56} & \textbf{85.90} \\
    28 & 93.04 & 73.07 & 81.85 & 83.80 \\
    56 & 93.23 & 70.73 & 80.44 & 82.80 \\
    \hline
  \end{tabular}
  \caption{Effect of shuffle granularity on model performance. We report Precision, Recall, F1, and Accuracy (\%). Best results are highlighted in bold.}
  \label{tab:shuffle_size_ablation}
\end{table}

\subsection{Effect of \texorpdfstring{$\alpha$}{alpha} in Contrastive Decoding}
\label{sec:ablation_alpha}

\begin{table}[t]
  \centering
  \small
  \setlength{\tabcolsep}{4pt}
  \begin{tabular}{lcccc}
    \hline
    \textbf{$\alpha$} & \textbf{Precision} & \textbf{Recall} & \textbf{F1} & \textbf{Accuracy} \\
    \hline
    0.0 & \textbf{94.74} & 73.27 & 82.63 & 84.60 \\
    0.4 & 94.55 & 75.20 & 83.77 & 85.43 \\
    0.8 & 93.54 & 76.20 & 83.98 & 85.47 \\
    1.2 & 93.50 & 76.67 & 84.25 & 85.67 \\
    1.6 & 93.00 & 77.00 & 84.25 & 85.60 \\
    2.0 & 92.68 & \textbf{77.67} & \textbf{84.51} & \textbf{85.77} \\
    \hline
  \end{tabular}
  \caption{Ablation on distillation weight $\alpha$. Results are reported on POPE (COCO / Random). The best result in each column is highlighted in bold (\%).}
  \label{tab:alpha_ablation_step04}
\end{table}

As shown in Table~\ref{tab:alpha_ablation_step04}, we investigate the impact of the distillation weight $\alpha$ in contrastive decoding on model performance. The experimental results indicate that the value of $\alpha$ plays a pivotal role in balancing Precision and Recall. As $\alpha$ increases from $0.0$ to $2.0$, the model exhibits a significant upward trend in Recall, improving from $73.27\%$ to $77.67\%$. Although this increase is accompanied by a slight decline in Precision (dropping from $94.74\%$ to $92.68\%$), the overall benefits outweigh the costs, as evidenced by the consistent improvements in both F1 score and Accuracy. Ultimately, with $\alpha=2.0$, the model achieves its peak performance on the POPE benchmark with an F1 score of $0.8451$ and an Accuracy of $0.8577$, demonstrating that this hyperparameter setting effectively enhances the comprehensive discriminative capability of the model.


\begin{table*}[t]
  \centering
  \small
  
  \setlength{\tabcolsep}{3.5pt}
  \renewcommand{\arraystretch}{1.05}
  \begin{tabular}{l|cccccccccc|c}
    \toprule
    \textbf{Decoding}
    & \textit{Exist.} & \textit{Count} & \textit{Pos.} & \textit{Color}
    & \textit{Posters} & \textit{Celeb.} & \textit{Scene}
    & \textit{Landmark} & \textit{Artwork} & \textit{OCR}
    & \textbf{Perception} \\
    \midrule
    Regular
    & 175.00 & \textbf{128.33} & 85.00 & \textbf{155.00}
    & 107.48 & 104.12 & 140.75 & 117.00 & 109.75 & \textbf{107.50}
    & 1229.93 \\
    VCD
    & 175.00 & 106.67 & 111.67 & 136.67
    & 123.13 & 135.88 & \textbf{142.50} & 145.00 & 113.00 & 102.50
    & 1292.01 \\
    SDCD
    & \textbf{190.00} & 126.67 & \textbf{125.00} & 143.33
    & \textbf{127.89} & \textbf{139.71} & 138.50 & \textbf{149.00} & \textbf{120.75} & 87.50
    & \textbf{1348.35} \\
    \bottomrule
  \end{tabular}
  \caption{Results on all MME perception-related tasks under the LLaVA backbone. The best performance for each column is bolded.}
  \label{tab:mme_perception_llava}

  \vspace{0.4cm} 

  \setlength{\tabcolsep}{6pt}
  \begin{tabular}{l|cccc|c}
    \toprule
    \textbf{Decoding}
    & \textit{Common Sense} & \textit{Numerical} & \textit{Text} & \textit{Code}
    & \textbf{Total} \\
    \midrule
    Regular
    & \textbf{117.14} & 52.50 & \textbf{77.50} & 60.00 & 307.14 \\
    VCD
    & 106.43 & 45.00 & 72.50 & 62.50 & 286.43 \\
    SDCD
    & 111.43 & \textbf{62.50} & 65.00 & \textbf{100.00} & \textbf{338.93} \\
    \bottomrule
  \end{tabular}
  \caption{Results on all MME cognition-related tasks under the LLaVA backbone. The best performance for each column is bolded.}
  \label{tab:mme_cognition_llava}

\end{table*}

\section{Detailed Experimental Results on MME}
\label{sec:detail_mme}

In Table~\ref{tab:mme_perception_llava}, we present the performance of different decoding strategies on the perception-related tasks of the MME benchmark. Experimental results indicate that SDCD exhibits consistent and significant performance advantages, achieving a total perception score of $1348.35$, which substantially outperforms both the Regular decoding strategy and the VCD baseline. This improvement is primarily attributed to SDCD's contrastive mechanism of structural enhancement and disruption. This mechanism effectively recalibrates the model's attention toward visual details, thereby enhancing the precise capture of image content while mitigating statistical biases.

Furthermore, Table~\ref{tab:mme_cognition_llava} shows the performance on cognition-related tasks. The results demonstrate that SDCD not only excels in perception tasks but also exhibits strong robustness in the cognitive dimension. Compared to the Regular decoding strategy and the VCD baseline, SDCD achieves significant improvements in tasks such as Numerical and Code, ultimately leading the baselines by a large margin with a total score of $338.93$. These findings provide compelling evidence that while enhancing visual perception capabilities, SDCD does not suffer from the performance fluctuations observed in certain baselines; instead, it further promotes the model's high-level reasoning abilities.

\end{document}